\documentclass[10pt,twocolumn,letterpaper]{article}

\usepackage{cvpr}
\usepackage{times}
\usepackage{epsfig}
\usepackage{graphicx}
\usepackage{amsmath}
\usepackage{amssymb}
\usepackage{algorithm}
\usepackage{algorithmic}
\usepackage{multirow}
\usepackage{booktabs}

\usepackage[pagebackref=true,breaklinks=true,letterpaper=true,colorlinks,bookmarks=false]{hyperref}
\usepackage[margin=4pt,font=footnotesize,labelfont=bf,labelsep=endash,tableposition=top]{caption}

\cvprfinalcopy

\def\sp{{ ~ }}

\def\bc{{\bf c}}

\def\bp{{\bf p}}

\def\bm{{\bf m}}

\def\0{{\bf 0}}
\def\1{{\bf 1}}

\def\bC{{\bf C}}

\def\bF{{\bf F}}

\def\bM{{\bf M}}
\def\bP{{\bf P}}

\def\mbR{{\mathbb R}}

\def\etal{\emph{et al. }}
\def\ie{\emph{i.e. }}
\def\eg{\emph{e.g. }}

\def\subsubsection{\noindent \textbf}
\def\bFins{{{\bf F}_\text{ins}}}
\def\bFseg{{{\bf F}_\text{seg}}}
\def\hbFins{{\widehat{\bf F}_\text{ins}}}
\def\hbFseg{{\widehat{\bf F}_\text{seg}}}

\begin{document}
	
	\title{Learning and Memorizing Representative Prototypes for 3D Point Cloud Semantic and Instance Segmentation}
	
	\author{
		Tong He $^{*}$         \sp
		Dong Gong \thanks{The first two authors contribute equally.}       \sp
		Zhi Tian        \sp
		Chunhua Shen \thanks{Corresponding author: $\tt chunhua.shen@adelaide.edu.au $.}    \sp
		\\
		The University of Adelaide 
	}
	\maketitle
	\thispagestyle{empty}

	\begin{abstract}

3D point cloud semantic and instance segmentation is crucial and fundamental for 3D scene understanding.
Due to the complex structure, point sets are distributed off balance and diversely, which appears as both category imbalance and pattern imbalance. As a result, deep networks can easily forget the non-dominant cases during the learning process, resulting in unsatisfactory performance. 
Although re-weighting can reduce the influence of the well-classified examples, they cannot handle the non-dominant patterns during the dynamic training. 
In this paper, we propose a memory-augmented network to learn and memorize the representative prototypes that cover diverse samples universally.
Specifically, a memory module is introduced to alleviate the forgetting issue by recording the patterns seen in mini-batch training.
The learned memory items consistently reflect the interpretable and meaningful information for both dominant and non-dominant categories and cases. 
The distorted observations and rare cases can thus be augmented by retrieving the stored prototypes, leading to better performances and generalization. 
Exhaustive experiments on the benchmarks, \ie S3DIS and ScanNetV2, reflect the superiority of our method on both effectiveness and efficiency. Not only the overall accuracy but also non-dominant classes have improved substantially.

\end{abstract}

	\section{Introduction}

\begin{figure}[htbp]
	\centering
	{
		\includegraphics[width=0.45\textwidth]{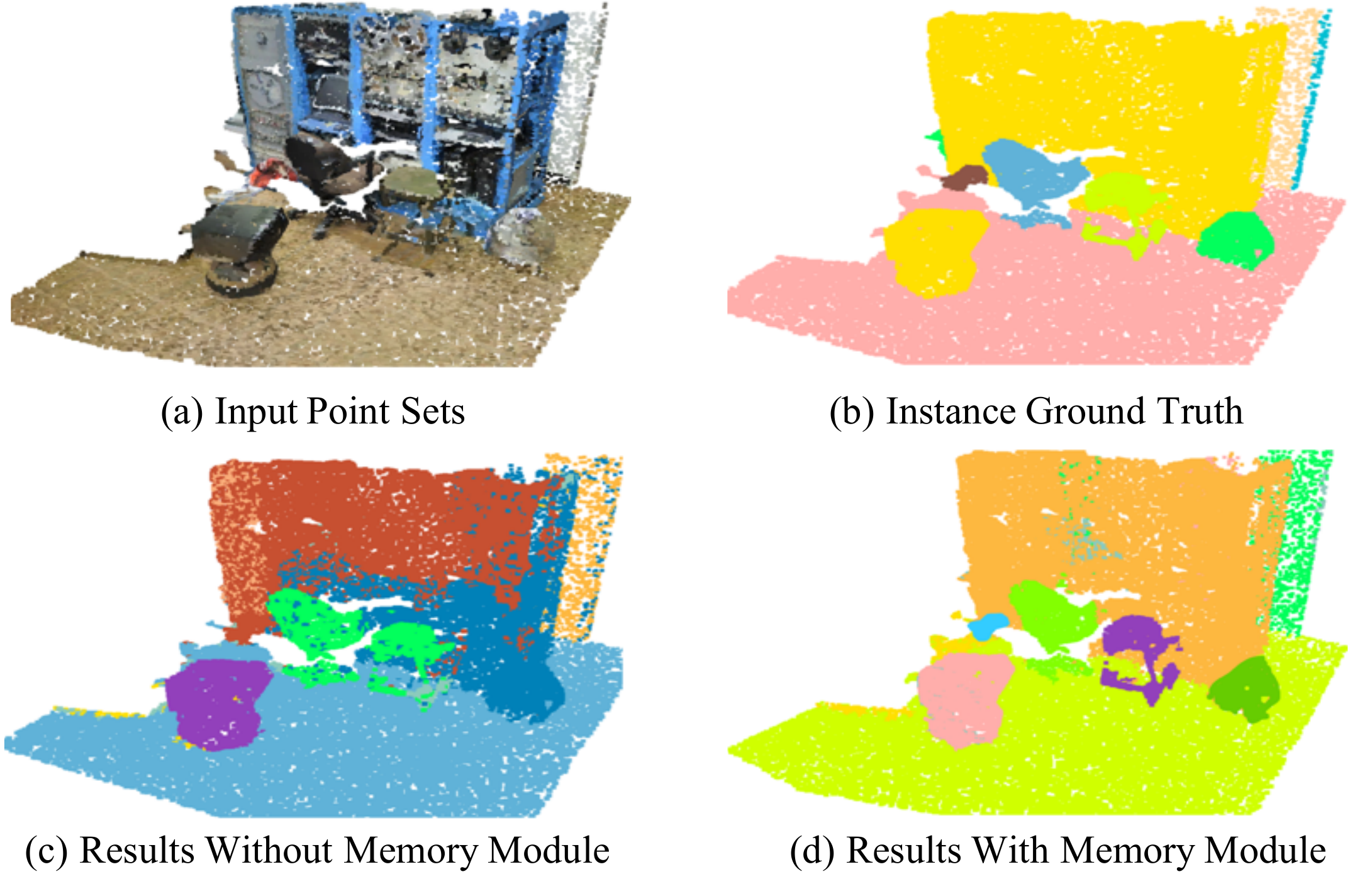}
		
	}
	\caption{Comparison of instance segmentation results between the proposed method with and without memory module. The performance of our method shows strong robustness against non-dominant cases.}
	\label{fig:mem_comp}
\end{figure}
The recent development of rapid and practical 3D sensors has provided easier ways to acquire 3D point cloud data, one of the widely used types of geometric data due to its simplicity \cite{qi2017pointnet}. 
3D scene understanding is critically important and fundamental for various applications, such as robotics, autonomous driving, and virtual reality. The core tasks include semantic segmentation and instance segmentation on point clouds, \ie assigning semantic labels and instance indication label for each point,  respectively. Comparing to the studies on 2D images \cite{he2018maskrcnn, bra2017cvprdis, dai2016instace}, semantic and instance on 3D point clouds lags far behind and have just started recently \cite{wang2018sgpn, wang2019asis, yi2018gspn, yang20193dbonet, hou20193dsis, jean20193dins}.

\par
Based on the pioneering works PointNet \cite{qi2017pointnet} and PointNet++ \cite{qi2017pointnetplusplus}, directly processing point sets becomes simpler, more memory-efficient and flexible than handling the volumetric grids with 3D convolution \cite{hou20193dsis, wu20153dshapenet, maturana2015voxnet}. 
Some following approaches \cite{wang2018sgpn, wang2019asis, yang20193dbonet, yi2018gspn} propose to handle semantic and instance segmentation in an end-to-end network jointly for fine-grained description of the scene. 
Specifically, discriminative instance embeddings are learned to measure the instance-level clustering patterns of the points \cite{wang2019asis, pham2019jsis3d}. 

Although existing methods have achieved some impressive results, we still can observe performance bottlenecks on different benchmark datasets \cite{armeni2016s3dis, dai2017scannet}, especially on the non-dominant classes with less samples (see Figure \ref{fig:comparison}). 
Suffering from the \emph{catastrophic forgetting} issue \cite{mccloskey1989catastrophic, toneva2018empirical}, deep networks can forget the non-dominant rare cases easily while learning on a dataset distributed off balance and diversely.
On point cloud data, imbalance issue usually appears as the \emph{category imbalance} and \emph{pattern imbalance}, which is severer than that on 2D images \cite{yang20193dbonet}. 
Firstly, discrepancy among the proportions of different categories are significant. In an indoor scene (see Figure \ref{fig:mem_comp} and \ref{fig:results}), most points belong to the background (\eg ground, ceiling, and wall), whereas the proportions of the objects (\eg chairs, desks and monitors) are much smaller. For example, in S3DIS \cite{armeni2016s3dis}, the total amount of ceiling points is 50 times larger than chair. 
Secondly, the patterns of the points are imbalanced and distributed diversely, which are often caused by the complex geometric informations, such as positions, shapes and relative relationships among instances. Some rare instance cases only have limited examples across the whole dataset.
For example, chairs are usually placed neatly in a conference room, while can also be placed in arbitrary positions (e.g., stacking and back-to-back) in an office room, as is shown in Figure \ref{fig:mem_comp}. 
Conventional methods \cite{yang20193dbonet} ignore this issue or simply resort to the focal loss \cite{lin2017focalloss}, by down weighing the well learned samples during training. 
However, they cannot directly handle the non-dominant patterns, which can be easily overwhelmed and forgotten. 

\par
\begin{figure*}[htbp]
	\centering
	{
		\includegraphics[width=0.9\textwidth]{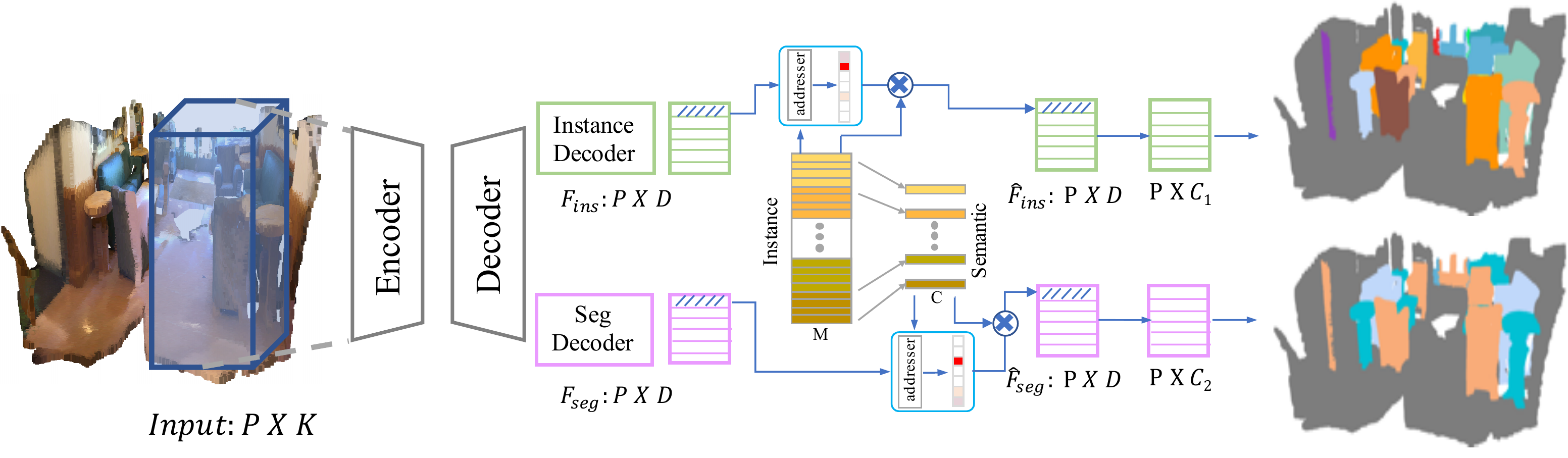}
		
	}
	\caption{The framework of our proposed MPNet, which contains two parallel branches with a shared encoder. A memory module is proposed to memorize representative prototypes that are shared by all samples. The maintained memory module is shared with all instances across different categories. Both distorted and rare cases can be augmented by retrieving the stored prototypes.}
	\label{fig:main}
\end{figure*}

To address the above issues, we propose to learn and memorize the discriminative and representative prototypes covering all the samples, which is implemented as a memory-augmented network, referred to as MPNet. 
The proposed MPNet includes two branches for predicting point-level semantic labels and obtaining per-point embedding for instance grouping, respectively.
As shown in Figure \ref{fig:main}, the two branches access a shared compact memory via two separate memory readers, which dynamically calibrate the per-point features with the memory items and associate the two tasks via the shared memory. 
Given an input, MPNet retrieves the most relevant items in the memory for the extracted per-point features and feeds only retrieved features to the following segmentation tasks.
Thus, driven by the task-specific training objectives, the compact memory is pushed to record the representative prototypes seen in mini-batches and associate them with newly seen patterns, alleviating the forgetting issues and strengthening the generalization. 

\par
In the proposed MPNet, the memory is maintained as a dictionary of the representative features and a semantic summarization, as shown in Figure \ref{fig:main}. Since the memory is trained to represent all the instances compactly, the learned prototypes can express a shared understanding of various instances. We observe that the learned memory items (\ie dictionary bases) can reflect interpretable and  meaningful informations, such as position and structure (see Figure \ref{fig:mem_vis}). 
Benefiting from the associative memory, the rare cases and distorted observations can be augmented by retrieving the stored prototypes, leading to better robustness and generalization. 
Additionally, different from previous methods relying on either pairwise relations computing \cite{wang2018sgpn} or KNN based feature aggregation \cite{wang2019asis}, the proposed MPNet is free from complex and time-consuming operations, which is more efficient. 

The main contributions are summarized as: 
\begin{itemize}
	\setlength{\itemsep}{0pt}
	\setlength{\parskip}{0pt}
	\setlength{\parsep}{0pt} 
	\item We propose a memory-augmented network for point cloud instance segmentation (\ie MPNet), which is trained to explicitly record the prototypes of the per-point features in a compact memory. The proposed MPNet is more effective and efficient than previous methods. 
	\item The learned prototypes can consistently represent interpretable and meaningful concepts of various instances, including dominant and non-dominant cases.
	\item Our proposed MPNet can boost the performance by a large margin with limited consumptions on computation and memory. State-of-the-art performance is achieved, showing the superiority on both effectiveness and efficiency.
\end{itemize}

	\section{Related Work}

\noindent \textbf{Deep Learning for 3D Point Cloud}~  
Existing methods for extracting features for 3D point cloud can be roughly categorized into three groups, including voxel-based \cite{wu20153dshapenet, maturana2015voxnet}, multi-view based \cite{dai20183dmv, hou20193dsis, qi2016multiview, su2015multiview} and point-based \cite{qi2017pointnet, qi2017pointnetplusplus, li2019deepgcns, qi2019deep, thomas2019kpconv}. 
\cite{maturana2015voxnet, wu20153dshapenet} are the pioneering works to transfer irregular points to regular volumetric grids, aiming to efficiently extract feature representation with 3D convolution. To reduce irrelevant operation on void places and save runtime memory usage, many works are proposed \cite{riegler2016octnet, graham2018sparseconv}.
Multi-view based methods extract features in both 2D and 3D domain. \cite{su2015multiview} is one of the pioneering multi-view based method, which apply view-pooling over the 2D predictions. 3D-SIS \cite{hou20193dsis}, proposed by Hou \etal, combine features from 2D and 3D via explicit spatial mapping in an end-to-end trainable network.
PointNet \cite{qi2017pointnet} is the first deep-learning-based work to operate directly on point sets, which uses shared MLP (multi-layer perceptron) to extract per-point feature. PointNet++ \cite{qi2017pointnetplusplus} improves the performance by extracting a hierarchical representation.
Many following works \cite{li2019deepgcns, wang2019graphatt, thomas2019kpconv, wu2018pointconv, li2018pointcnn} have been proposed to get a better representation of local context. 
Due to its simplicity, we select PointNet++ as our backbone and leave the choices of other backbones for future work.

\noindent \textbf{Instance Segmentation on Point Cloud}~ 
Deep-learning-based instance segmentation for 3D point cloud is rarely studied until huge application potential has been discovered recently. 
SGPN \cite{wang2018sgpn} is the first deep learning based method working on this field. It first splits the whole scene into separate blocks. For every single block, per-point grouping candidates are proposed by predicting a similarity matrix that reflects affinity between each pair of points. A block merging algorithm is conducted for post-processing by taking segmentation results of the overlapped area into consideration.
However, huge memory is needed for storing the pair-wise matrix, which makes it memory-consuming for post-processing. In order to solve this, 
Wang \etal proposed ASIS \cite{wang2019asis}, which utilized a discriminative loss function \cite{bra2017cvprdis} to encourage points belonging to the same instance are mapped to a metric space with close distances. Moreover, in order to make the two tasks take advantage of each other, convolution and KNN search are applied for mutual feature aggregation of the two tasks, making it inefficient and time-consuming. 

\noindent \textbf{Memory Networks}~
Memory based approaches have been discussed for solving various problems. NTM \cite{graves14ntm} is proposed to improve the generalization ability of the network by introducing an attention-based memory module.  
Gong \etal \cite{gong2019memorizing} proposed a memory augmented auto-encoder for detecting anomaly. Anomaly is detected by represented the input with prototypical elements of the normal data maintained in a memory module. Prototypical Network \cite{jake2017nips} maintains a category-wise templates for the problem of few-shot classification. Liu \cite{liu2019longtail} proposed an OLTR algorithm to solve the open-ended and long-tail problem by associating a memory feature that can be transfered to both head and tail classes adaptively.

	\section{The Proposed Method}

\subsection{Overview of the Proposed MPNet}

We propose to tackle the imbalance issue in point cloud semantic and instance segmentation by learning and memorizing prototypes of the cases seen during training. The discriminative and representative prototypes are stored in a memory module and can be accessed via specific readers.
As shown in Figure \ref{fig:main}, the proposed memory-augmented network (\ie MPNet) adopts an encoder-decoder architecture, which is free from the specific design of the encoder and decoder. 
In the proposed MPNet, we use PointNet++ \cite{qi2017pointnetplusplus} to implement the encoder for per-point feature extraction. Two parallel decoders for instance segmentation and semantic segmentation are built upon the shared encoder. As described in the following, the memory is implemented as a dictionary to record the prototypes as bases. 
For the both branches, given a per-point feature, two specifically designed memory readers are applied to generate addressing weights to access the memory, respectively, via soft-attention. 
The retrieved items from the memory are then applied for the following semantic labeling and instance grouping tasks. 
For any input sample, the relevant memory items are retrieved for the two tasks and also updated via prorogations driven by the task objectives and specifically designed instance regularizer (described in Section \ref{sec:reg}).

\par
Given a set of input points $\{\bp_i\}_{i=1}^P$ with $\bp_i\in \mbR^K$, we can formulate the input of the network as a matrix $\bP\in \mbR^{P\times K}$, where $K$ denotes the input feature dimension and $P$ denotes the total number of input points. Input features of each points may consist of both geometry and appearance information, \ie 3D coordinate $(x,y,z)$ and RGB values. The two branches produces features $\bF_\text{seg}\in \mbR^{N\times D}$ and $\bF_\text{ins}\in \mbR^{N\times D}$, respectively, where $D$ denotes the dimension of features. 
Instead of directly using $\bF_\text{seg}$ and $\bF_\text{ins}$ to perform semantic and instance segmentation tasks, respectively, MPNet applies them as queries to retrieve the prototypes in the memory and then obtain alternative features $\hbFseg$ and $\hbFins$, which are delivered to the following semantic classifier and instance embedding module. The memory is randomly initialized and updated during training. The two branches access the memory with specifically designed read heads,

\subsection{Memory Representation for Prototypes}
The \emph{prototypes memory} is designed as a matrix $\bM \in \mbR^{N\times D}$, where $N$ is a hyper-parameter that defines the number of memory slots and $D$ is the feature dimension that is identical with the outputs from the two branches. The $N$ memory slots are used to restore the prototypes shared by all the instances across all categories.
To easily represent the semantic characteristics, we define a \emph{semantic memory} $\bC\in \mbR^{C\times D}$ of $\bM$, where $C$ denotes the number of categories for the semantic segmentation task and each row of $\bC$ represents the summary of a class.
Although the memory slots in prototypes memory $\bM$ are shared to represent universal concepts of the all instances, to generate semantic summary $\bC$ from $\bM$, we equally associate the $N$ memory slots in $\bM$ with $C$ categories and thus define $N=N_c\times C$, where $N_c$ is denoted as the number per-category prototypes.
As shown in Figure \ref{fig:main}, the $i$-th row in $\bC$, \ie $\bc_i$, can be seen as a average of the $i$-th subsegment in $\bM$, \ie rows in $\bM$ from $(i-1)\times N_c+1$ to $i\times N_c$. 
Specifically, we obtain $\bc_i$ by averaging the submatrix $\bM_i$:
\begin{equation}
\bc_i = \frac{1}{N_c} \sum\nolimits_{j=(i-1)\times N_c+1}^{i\times N_c} \bm_j,
\end{equation}
where $\bm_j$ denotes the $j$-th row vector of $\bM$. 
 
\par
Given the query features $\bFins$ and $\bFseg$, the instance grouping branch directly addresses the prototypes memory $\bM$ and the semantic labeling branch accesses the semantic summary $\bC$, with two specifically designed readers.
$\bM$ can be seen as an dictionary to restore the representative bases shared by all instances, since the instances cross different categories can share some common basic components and characteristics.
As the semantic memory $\bC$ is a re-parameterization of $\bM$, the two tasks are naturally associated together, without computation-consuming operations as \cite{wang2019asis}.
Due to the supervision from both tasks, the learned and memorized prototypes are discriminative not only for grouping instances but also for semantic classification.

\begin{figure}[htbp]
	\centering
	{
		\includegraphics[width=8cm, height=4cm]{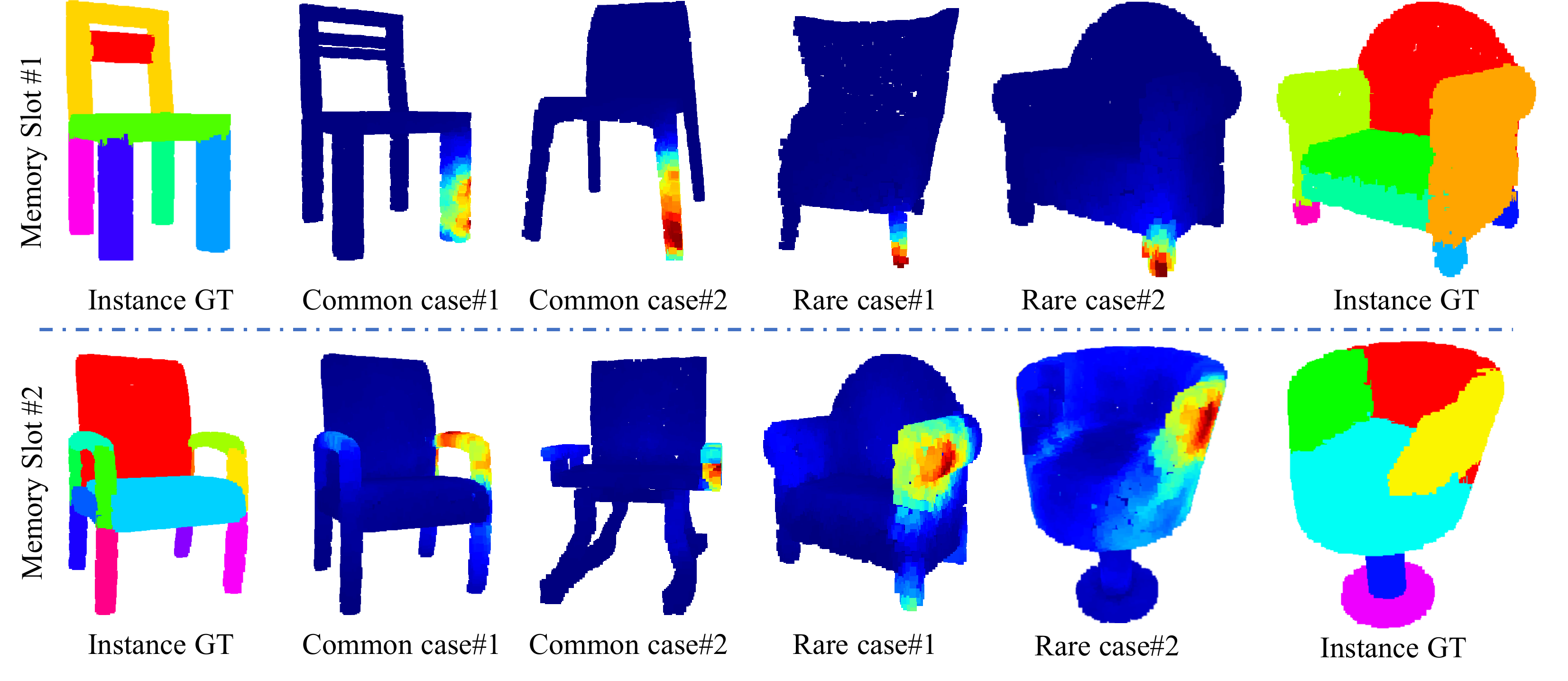}
		
	}
	\caption{Visualization of the memory representation. The goal is to find out part instance segmentation within an object, \ie four chair legs in each example are four different instances. Even with various external shapes appearances, the instance memory successfully capture consistent representation for both dominant cases and rare cases.}
	\label{fig:mem_vis}
\end{figure}

\subsection{Memory-augmented Instance Embedding}
\label{sec:reg}
\noindent \textbf{Memory reader for instance segmentation}
Given the $i$-th instance query feature $\textbf{f}_{\text{ins}, i}$ from $\bF_{ins}$, an attention-based reader is proposed to address the most relevant prototypes from $\bM$. The soft addressing weights $w$ is calculated as follows:
\begin{equation}
w_{ij} = \frac{exp(d(\textbf{f}_{\text{ins}, i}, \bm_j))}{\sum_{j=1}^{N}exp(d(\textbf{f}_{\text{ins}, i}, \bm_j))},
\label{eq:w}
\end{equation}
where $\bm_j$ is the $j$-th row vector of $\bM$ and $d(\cdot, \cdot)$ is function for measuring similarity of the $i$-th query item and the $j$-th prototype item. In MPNet, we utilize cosine distance. The alternated $i$-th instance feature $\widehat{\bf f}_{\text{ins}, i}$ from $\hbFins$ can be calculated through: $\widehat{\bf f}_{\text{ins}, i} = \sum_{j=1}^{N}w_{ij} \bm_j$.

To have a better understanding of the learned memory prototypes, we select the category of `Chair' in PartNet \cite{mo2019partnet} for training and visualization, as shown in Figure \ref{fig:mem_vis}. Each chair is a testing sample and the goal is to find out instance part of object. For example, the four chair legs from a chair are noted as different instances. We select two memory prototypes $\bm_i$ and $\bm_j$, and different colors in a chair refer to the addressing weights of the $i$-th and $j$-th columns in w (see Eq. (\ref{eq:w})).  For each memory item, the points that are addressing it have consistent geometric meaning, as shown in Figure \ref{fig:mem_vis}. 
The consistency of the learned prototypes allows it to capture discriminative representation for both dominant and rare cases.

\subsubsection{Instance-aware regularization}
The prototypes memory are updated via propagation driven by the training objectives. To make it effective, specifically designed regularization term $R_{\text{ins}}$ is proposed, defined as follows:
\begin{equation}
R_{\text{ins}} = \frac{1}{K} \sum_{k=1}^{K} \frac{1}{N_k}\sum_{n=1}^{N_k} \|G(\widehat{\bf f}_{\text{ins}, n}) - GT_k \| ^2,
\label{eq: reg_i}
\end{equation}
where $K$ is the instance number, $N_k$ is the point number of $k$-th instance, $G(\cdot)$ is a simple MLP that predicts geometric centroid of the $k$-th instance. $GT_k$ is the corresponding geometrical ground truth. 

\subsection{Memory-augmented Semantic Labeling}
\subsubsection{Memory reader for semantic segmentation}
Similar to the instance reader, the segmentation reader outputs an addressing weights for retrieving the most relevant categories. Given the $i$-th segmentation query $\bf f_{\text{seg}, i}$  from $\bFseg$ and $j$-th memory item $\bc_j$, the feature for semantic segmentation is calibrated by: $\widehat{\bf f}_{\text{seg}, i} = \alpha^{T} \bC  =  \sum_{j=1}^{C}\alpha_{ij} \bc_j$.
where $\bc_j$ is the $j$-th centroid for segmentation and $\alpha_{ij}$ is the similarity coefficients between $\bf f_{\text{seg}, i}$ and $\bc_j$, similar to Eq. (\ref{eq:w}).

\subsubsection{Semantic memory regularization}
To force the centroids of different classes, \ie the semantic summarization $\bC$, to keep a separable distance,  $R_{\text{seg}}$ is proposed to regularize the large margin of the inter-class and the compactness of inner-class. Given the $i$-th calibrated feature $\widehat{\bf f}_{\text{seg}, i}$ and its semantic label $y_i$, the regularization term $R_{\text{seg}}$ is calculated as:
\begin{equation}
\begin{aligned}
R_{\text{seg}}= \text{max}(0, &\sum_{j=y_i}\| \widehat{\bf f}_{\text{seg}, i} - \bc_j \| \\
 &- \sum_{j\neq y_i} \| \widehat{\bf f}_{\text{seg}, i} - \bc_j  \| + m ),
\end{aligned}
\label{eq: reg_s}
\end{equation}
where $m$ is the relaxation margin, which is set to 5 in all our experiments. Each $\bc_j$ performs like an anchor point and pull the features with identical semantic labels close to it and push the features with different semantic labels away from it. 

\subsection{Loss Functions}
\subsubsection{Classification loss}
We use cross entropy loss $L_{\text{CE}}$ for the semantic segmentation task. Instead of using softmax for normalization, we found squashing function proposed by \cite{sara2017nips, liu2019longtail} provides a little higher performance on the accuracy of semantic segmentation. To summarize, classification loss is defined as:
\begin{equation}
L_{\text{CE}} = \frac{1}{P}\sum_{n=1}^{P} \text{CE}(\frac{\| fc(\widehat{\bf f}_{\text{seg}, i}) \|^2}{1+\| fc(\widehat{\bf f}_{\text{seg}, i}) \|^2} \cdot \frac{fc(\widehat{\bf f}_{\text{seg}, i})}{\| fc(\widehat{\bf f}_{\text{seg}, i}) \|},  y_n),
\end{equation}
where $\text{CE}$ refers to the cross entropy loss and $P$ is the total number of examples. $fc(\cdot)$ is a fully convolution operation that projects the calibrated $\widehat{\bf f}_{\text{seg}, i}$ to the classification space.

\subsubsection{Instance discriminative loss}
\label{section:ins}
Similar to \cite{wang2019asis}, given the retrieved instance features $\{\widehat{\bf f}_{\text{ins}, i}\}_{i=1}^{P}$, a simple layer perceptron is utilized to project the feature to the embedding space $\{\textbf{e}_{\text{ins},i}\in \mathbb{R}^{c'}\}_{i=1}^{P} $ with feature dimension $c'$ (we set $c'=5$ in all our experiments). The loss is formulated as follows:
\begin{equation}
\begin{aligned}
L_{\text{dis}} &= \frac{1}{K} \sum_{k=1}^{K} \frac{1}{N_k} \sum_{n=1}^{N_k} \left[ \| \textbf{e}_{\text{ins},n} - \boldsymbol{\mu}_k  \| -{\sigma}_v \right]_{+}^2 \\
&+ \frac{1}{K(K-1)}\mathop{\sum_{i = 1}^{K} \sum_{j = 1}^{K}}_{i \neq j}\left[ 2{\sigma}_d - \| \boldsymbol{\mu}_i - \boldsymbol{\mu}_j \| \right]_+^2,
\label{eq:dis_loss}
\end{aligned}
\end{equation}
where $K$ is the instance amount and $N_k$ is the number of $k$-th instance and $\boldsymbol{\mu}_k$ is the average embedding of the $k$-th instance, which is calculated by $ \boldsymbol{\mu}_k = \frac{1}{N_k} \sum_{n=1}^{N_k}\textbf{e}_{\text{ins},n}$. 
$\sigma_v$ and $\sigma_d$ in Eq. \eqref{eq:dis_loss} are respectively the margins for the variance and distance loss terms as defined in \cite{bra2017cvprdis, wang2019asis}.

\subsubsection{Training objective}
As all operations are differentiable, prototypes memory module can be updated through back-propagation in an end-to-end manner. By combining the four losses discussed above, the training objective is formulated as: 
\begin{equation}
L = L_{\text{CE}} + L_{\text{dis}} + R_{\text{seg}} + \lambda R_{\text{ins}},
\end{equation}
we found $R_{\text{ins}}$ is sensitive to the learning rate and we set it 0.1 and maintain the others to 1.0 in all our experiments.

	\section{Experiments}

To validate the effectiveness of our proposed method, both qualitative and quantitative experiments are conducted on two public datasets: Stanford 3D Indoor Semantic Dataset (S3DIS) \cite{armeni2016s3dis} and ScanNetV2 \cite{dai2017scannet}. 
\subsection{Datasets}
S3DIS dataset \cite{armeni2016s3dis} covers more than 6000 $m^2$ and is collected in 6 large-scale indoor areas. It includes 272 rooms and more than 215million points, each of which contains both instance and semantic annotations out of 13 classes. ScanNetV2 is another large-scale dataset for point cloud instance segmentation, which consists of 1613 indoor scans from 40 categories. The dataset is split into 1201, 312 and 100 for training, validating and testing, respectively.

\subsection{Evaluation}
Following \cite{wang2019asis} on the S3DIS dataset, the performance on Area-5 and k-fold cross-validation are reported in our experiments. For semantic segmentation, we present the overall accuracy (oAcc), which measures point-level accuracy, mean class accuracy (mAcc), which calculates average category-level accuracy and mean intersection-over-union (mIoU), which provides a measure for matching predicted segmentation results and the ground truth across all categories. 
For instance segmentation, four evaluation metrics are calculated, namely, $mConv$, $mWConv$, $mPrec$ and $mRec$. $mConv$  is defined as the mean instance-wise matching IoU score between ground truth and prediction. Instead of treating every instance equally, $mWConv$ is weighted by the size of each instance object. Moreover, traditional $mPrec$ and $mRec$ represents mean precision and mean recall with IoU threshold 0.5, which are widely used in the 2D image object detection task.

\subsection{Implementation Details}
For the S3DIS and ScanNetV2, similar to PointNet \cite{qi2017pointnet}, each room is divided into $1m \times 1m$ blocks with a stride of $0.5m$. 4096 points are randomly sampled as input from each block during the training process. 
The feature for each point is consist of both color and geometric information, \ie $R, G, B, X, Y, Z \dots$. Without special notation, all experiments are conducted using vanilla PointNet++ \cite{qi2017pointnetplusplus} as backbone (without introducing any multi-scale grouping operation). We use ADAM optimizer with initial learning rate of 1e-2, momentum of 0.9 and batch size of 16. The learning rate is divided by 2 for every $3 \times 10^{5}$ iterations.
The hyper-parameters for metric learning are selected to be the same with \cite{wang2019asis}, namely, $\sigma_v = 0.5$, $\sigma_d = 1.5$. The number for memory slots is set to 150 per-category. The loss weight for $L_{}$ is set to 0.01, which has a significant influence to the final performance. 
The whole network is trained end-to-end for 100 epochs in total. 

During inference time, blocks within each room are merged in a snack pattern by utilizing the segmentation and instance results of the overlapped region. Detailed settings of the algorithm are identical with \cite{wang2018sgpn}. 

\subsection{Ablation Study}
In this section, we describe the influence of each integration of aforementioned components. All the results are tested on S3DIS Area-5 for fair comparison. 
We first build a strong baseline which is similar to ASIS vanilla \cite{wang2019asis}. The framework has two independent decoders which are responsible for semantic segmentation and instance metric grouping, respectively. Using Pointnet++ as backbone, the baseline model achieves 52.3 $mPrec$ and 41.4 $mRec$ on S3DIS Area-5, which is 16.3 and 12.7 higher than SGPN \cite{wang2018sgpn}, respectively. 
Built upon the strong baseline, our MPNet surpass it by a large margin via  memorizing representative prototypes. In the following section, provide detailed analysis on different parts.

\begin{table}[!t]
	\caption{Ablation study on the S3DIS dataset Area-5 set with vanilla Pointnet++ as backbone. \textbf{FL} refers to focal loss. \textbf{InsMem} means the memory is updated by instance information. \textbf{SegMem} means the memory is updated by semantic segmentation supervision. \textbf{Regul} refers to the regularizations used in learning the prototypes memory. Both instance segmentation and semantic segmentation results are provided. }
	\small 
	\centering
	\addtolength{\tabcolsep}{-3.0pt}
	\begin{center}
		\begin{tabular}{c|cccc|c|c|c}
			\toprule[0.2 em]
			
			Method & FL  & InsMem & SegMem  &Regul & mPre &mRec &oAcc  \\
			\toprule[0.2 em]
			Baseline & &  & & &52.3 &41.4 &86.2 \\
			&\checkmark  & & & &55.2 &43.0 &86.9 \\
			\toprule[0.1 em]
			& &\checkmark & & &58.9 &47.0 &87.7 \\
			& &\checkmark &\checkmark & &60.2 &47.2 &88.1 \\
			Ours& &\checkmark &\checkmark &\checkmark &62.5 &49.0 &88.2 \\
			
			\bottomrule[0.1 em]
		\end{tabular}
	\end{center}
	\vspace{-1.5em}
	\label{tab:s3dis_ablation_results}
\end{table}

\subsubsection{Focal Loss.} 
The discrepancy among different categories are significant in 3D point cloud.
Focal loss \cite{lin2017focalloss} has been widely used in different kinds of vision tasks due to the imbalance of data distribution. It addresses the problem by down-weighting the well-classified samples. However, it only alleviates the category imbalance to some extent and fail to solve the diverse distributed patterns. 
As shown in Table \ref{tab:s3dis_ablation_results}, focal loss can only improve the mean precision and mean recall by 2.9 and 2.4, respectively. 
Compared with Focal Loss, our method is more powerful to solve both data imbalance and pattern imbalance by recording and memorizing the prototypical patterns. 

\begin{figure}[htbp]
	\centering
	{
		\includegraphics[width=0.45\textwidth]{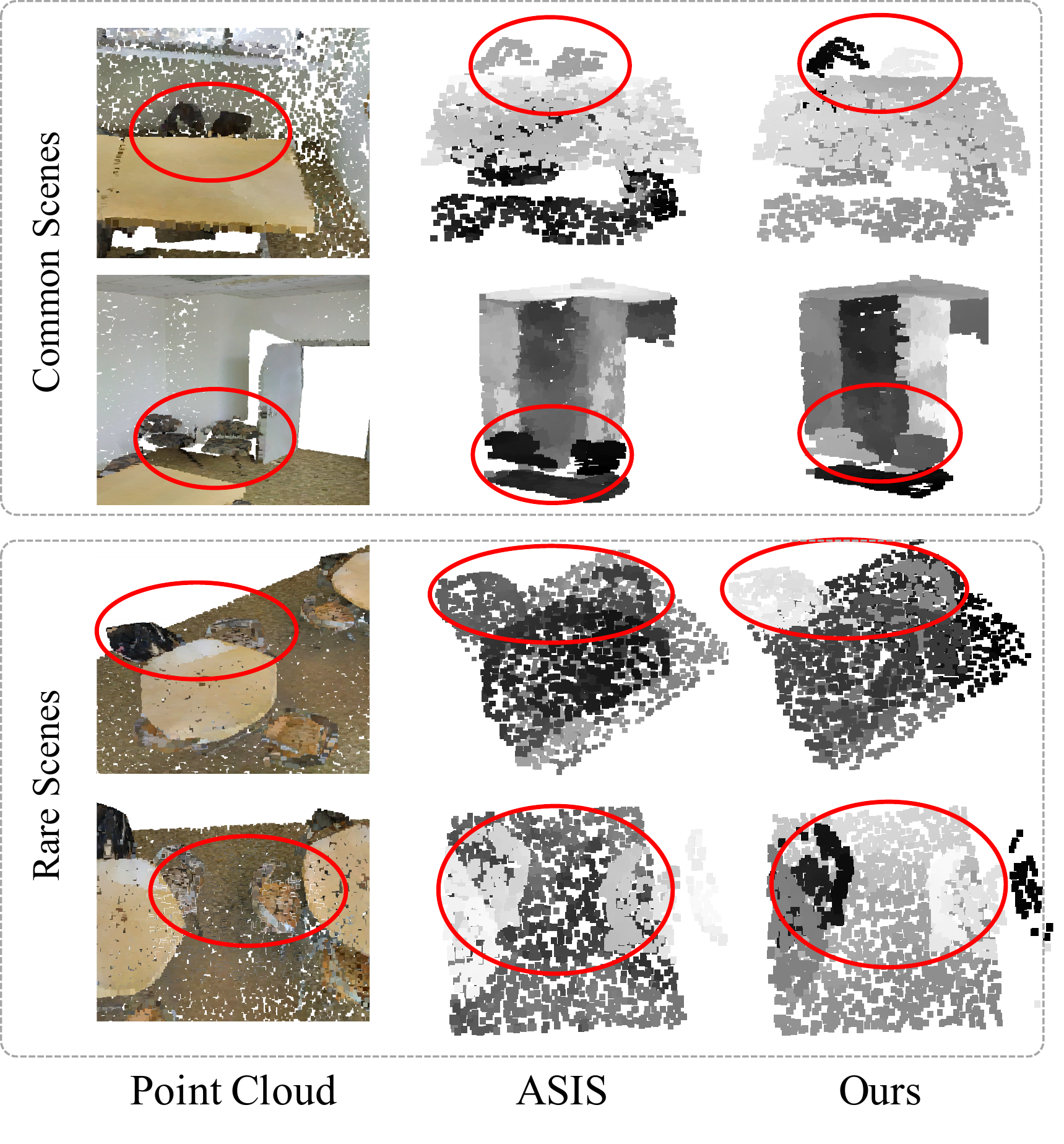}
		
	}
	\caption{Barnes-Hut t-SNE \cite{vandermaaten14a} visualization of our instance embedding on S3DIS Area-5 set (Best viewed when zoom in). The embedding feature is projected to 1-D and the distances is normalized to unit length so that the gap of gray-scale between different instances reflects the distances in the embedding space.}
	\label{fig:tsne}
\end{figure}

\begin{table}[ht]
	\caption{Instance Segmentation results on S3DIS dataset. Both Area-5 and 6-fold results are reported. All our results are achieved based on a vanilla PointNet++ backbone (without multi-scale grouping) for fair comparison.}
	\begin{center}
		\small 
		\setlength{\tabcolsep}{3.8pt}
		\begin{tabular}{c|c|c|c|c|c}
			\hline 
			\hline
			Method &Year   & mCov    & mWCov    &  mPrec & mRec \\
			\hline
			\hline
			\multicolumn{6}{c}{Test on Area 5} \\
			\hline
			SGPN ~\cite{wang2018sgpn} &2018 &  32.7  & 35.5  &  36.0  & 28.7   \\
			
			ASIS ~\cite{wang2019asis} &2019 & 44.6  &  47.8 &  55.3 &  42.4 \\
			3D-BoNet ~\cite{yang20193dbonet} &2019 & -  &  - &  57.5 &  40.2 \\
			
			\textbf{Ours}    &-  & \textbf{50.1} &\textbf{53.2} &\textbf{62.5} &\textbf{49.0} \\

			\hline 
			\hline
			\multicolumn{6}{c}{Test on 6-fold} \\
			\hline
			SGPN ~\cite{wang2018sgpn} &2018 &  37.9  & 40.8  &  31.2  & 38.2   \\
			MT-PNet ~\cite{pham2019jsis3d} &2019 &-   &-   &24.9 &- \\
			MV-CRF ~\cite{pham2019jsis3d} &2019 &- &- &36.3 &- \\
			ASIS ~\cite{wang2019asis} &2019 &51.2 &55.1  & 63.6 & 47.5 \\
			3D-BoNet ~\cite{yang20193dbonet} &2019 &- &- &65.6 &47.6 \\
			PartNet ~\cite{mo2019partnet} &2019 &- &- &56.4 &43.4 \\ 
			\textbf{Ours}        &-   &\textbf{55.8} &\textbf{59.7} &\textbf{68.4} &\textbf{53.7} \\
			\hline
		\end{tabular}
	\end{center}
	
	\label{tab:s3dis_ins_results}
\end{table}

\begin{table*}[htb]
	\caption{Comparison per-class performance of our proposed method with state-of-the-art on S3DIS semantic segmentation task, tested on all areas.  Our result utilize vanilla pointnet++ \cite{qi2017pointnetplusplus} without multi-scale group. Even with a simple baseline, the proposed method surpassed the graph based method by more than 1\% with mIOU.}
	\begin{center}
		\footnotesize 
		\setlength{\tabcolsep}{4.2pt} 
		\begin{tabular}{lcc|ccccccccccccc}
			\toprule
			\textbf{Method}  & \textbf{OA} & \textbf{mIOU}    & \textbf{ceiling} & \textbf{floor}   & \textbf{wall}    & \textbf{beam}    & \textbf{column}  & \textbf{window}  & \textbf{door}    & \textbf{table}   & \textbf{chair}   & \textbf{sofa}    & \textbf{bookcase} & \textbf{board}   & \textbf{clutter} \\
			\midrule
			PointNet ~\cite{qi2017pointnet}& 78.5    & 47.6    & 88.0    & 88.7    & 69.3    & 42.4    & 23.1    & 47.5    & 51.6    & 54.1    & 42.0    & 9.6     & 38.2    & 29.4    & 35.2    \\
			MS+CU ~\cite{engelmann2017iccvw}   & 79.2    & 47.8    & 88.6    & \textbf{95.8} & 67.3    & 36.9    & 24.9    & 48.6    & 52.3    & 51.9    & 45.1    & 10.6    & 36.8    & 24.7    & 37.5    \\
			G+RCU ~\cite{engelmann2017iccvw}    & 81.1    & 49.7    & 90.3    & 92.1    & 67.9    & 44.7 & 24.2    & 52.3    & 51.2    & 58.1    & 47.4    & 6.9     & 39.0    & 30.0    & 41.9    \\
			PointNet++ ~\cite {qi2017pointnetplusplus} &-  &53.2 &90.2 &91.7 &73.1 &42.7 &21.2 &49.7 &42.3 &62.7 &59.0 &19.6 &45.8 &48.2 &45.6\\
			PointNeighbor ~\cite{engelmann2018} &- & 58.3 &92.1 &90.4 &\textbf{78.5} & 37.8 &35.7 & 51.2 & 65.4 & 64.0 & 61.6 &25.6 & 51.6 & 49.9 & 53.7 \\
			DGCNN ~\cite{wang2019dgcnn}  & 84.1    & 56.1    & - & - & - & - & - & - & - & - & - & - & - & - & - \\
			ResGCN-28 ~\cite{li2019deepgcns}  & 85.9    & 60.0 & 93.1 & 95.3    & 78.2 & 33.9    & \textbf{37.4} & 56.1 & \textbf{68.2} & 64.9 & 61.0  & 34.6    & 51.5    & 51.1 & \textbf{54.4}\\
			Ours PointNet++ &\textbf{86.8} &\textbf{61.3} &\textbf{94.0} &94.1 &76.6 & \textbf{53.4} & 33.6 & 54.2 & 62.7 &\textbf{70.2} & 60.2 &\textbf{36.6} &53.4 &\textbf{54.3} &53.5 \\
			\bottomrule
		\end{tabular}
	\end{center}
	
	\label{tab:seg_s3dis}
\end{table*}
\begin{figure}[htbp]
	\centering
	{
		\includegraphics[width=0.45\textwidth]{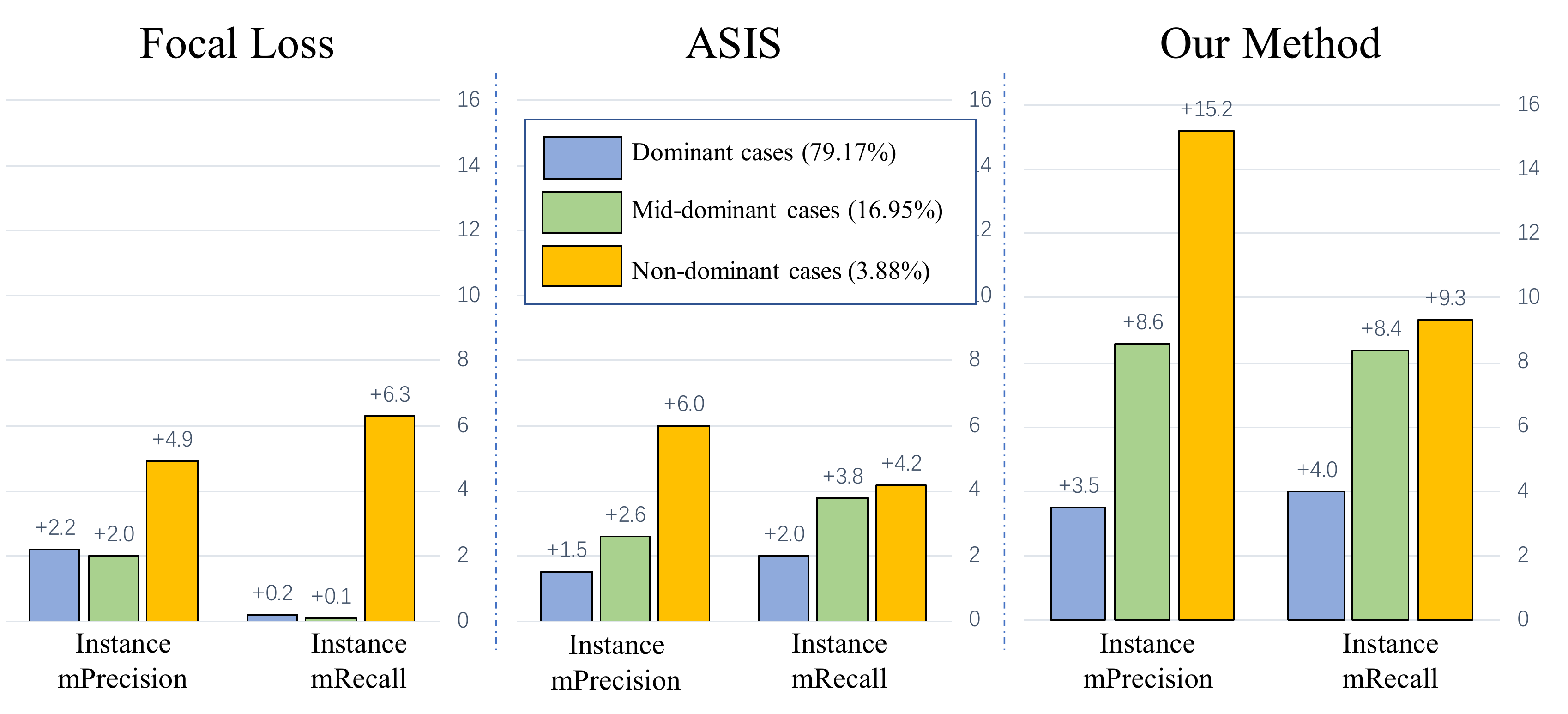}
	}
	\caption{The comparison of improvements between our proposed method and baseline model with focal loss \cite{lin2017focalloss} and ASIS \cite{wang2019asis}. Both mean precision and mean recall of instance are reported.
	}
	\label{fig:comparison}
\end{figure}

\subsubsection{Prototypes Memory M and C.}
The representative and consistent prototypes are maintained in a memory module \textbf{M}, which is shared to represent universal concepts of all instances. Besides, a semantic memory \textbf{C} is served as a prototypes summary to efficiently represent the semantic characteristics. As shown in Table \ref{tab:s3dis_ablation_results}, using instance memory \textbf{M} alone can boost $mPre$ from 52.3\% to 58.9\% and $mRec$ from 41.4\% to 47.0\%. 
On the other hand, using segmentation memory $C$ can bring another 1.3\% and 0.5\% improvement with the metric of $mPrec$ and $oAcc$. As two tasks are highly correlated due to the shared encoder backbone, utilizing \textbf{M} can also brings about 1.5\% improvement for semantic segmentation in terms of $oAcc$.

\subsubsection{Regularization Loss.} 
To effectively learn representative and discriminative prototypes, regularization losses are proposed in Eq. \eqref{eq: reg_s} and Eq. \eqref{eq: reg_i}. The first one is to keep large-margin between different categories from memory \textbf{C}. The second one is designed for forcing the calibrated instance embeddings to have identical geometric output. As shown in Table \ref{tab:s3dis_ablation_results}, these regularizations can boost the $mPre$ and $mRec$ for about $1.7$ and $1.8$, respectively. 
\subsubsection{The Impact of Memory Size.} We study the influence of the memory size to the final performance. We set three values of $N_c$ with 100, 150, 200 as the number of per-category prototypes. The $mPrec$ on S3DIS Area-5 are 60.4, 62.7 and 62.5 respectively. 
The results show that the performance increases as $N_c$ grows, and become stable after 200. In all our experiments, $N_c$ is set to 150.

\subsubsection{Visualization of the Memory Representation.} 
Given the input features, the most relevant prototypes are retrieved to calibrate the features. 
In Figure \ref{fig:tsne}, we visualize the embedding features with and without the memory module.
Both common and rare scenes, \ie office and lobby, are selected, according to the amount of training samples. The embeddings are projected to 1-D with the help pf Barnes-Hut t-SNE \cite{vandermaaten14a}. In both situations, our MPNet generate more discriminative embedding features, which is critical for separate different instance.

\subsection{Comparison with the State-of-the-art}
\subsubsection{Performance on non-dominant cases.}
We first compare the performance of our proposed MPNet with state-of-the-art method ASIS \cite{wang2019asis} on non-dominant cases. We first sort the 13 categories on S3DIS according to the total amount of training samples, and split the dataset into three levels: dominant cases (the first 4 classes), mid-dominant cases (the mid 5 classes) and non-dominant cases (the last 4 classes). The amount proportions of the three levels are 79.17\%, 16.95\% and 3.88\%, respectively. As shown in Figure \ref{fig:comparison}, we report the improvement with two metrics: $mPrec$, $mRec$. Our method can not only boost the performance on dominant cases, but surpass focal loss and ASIS \cite{wang2019asis} by a large margin on non-dominant cases.

\begin{figure*}[htbp]
	\begin{center}
	{
		\includegraphics[width=1.0\textwidth]{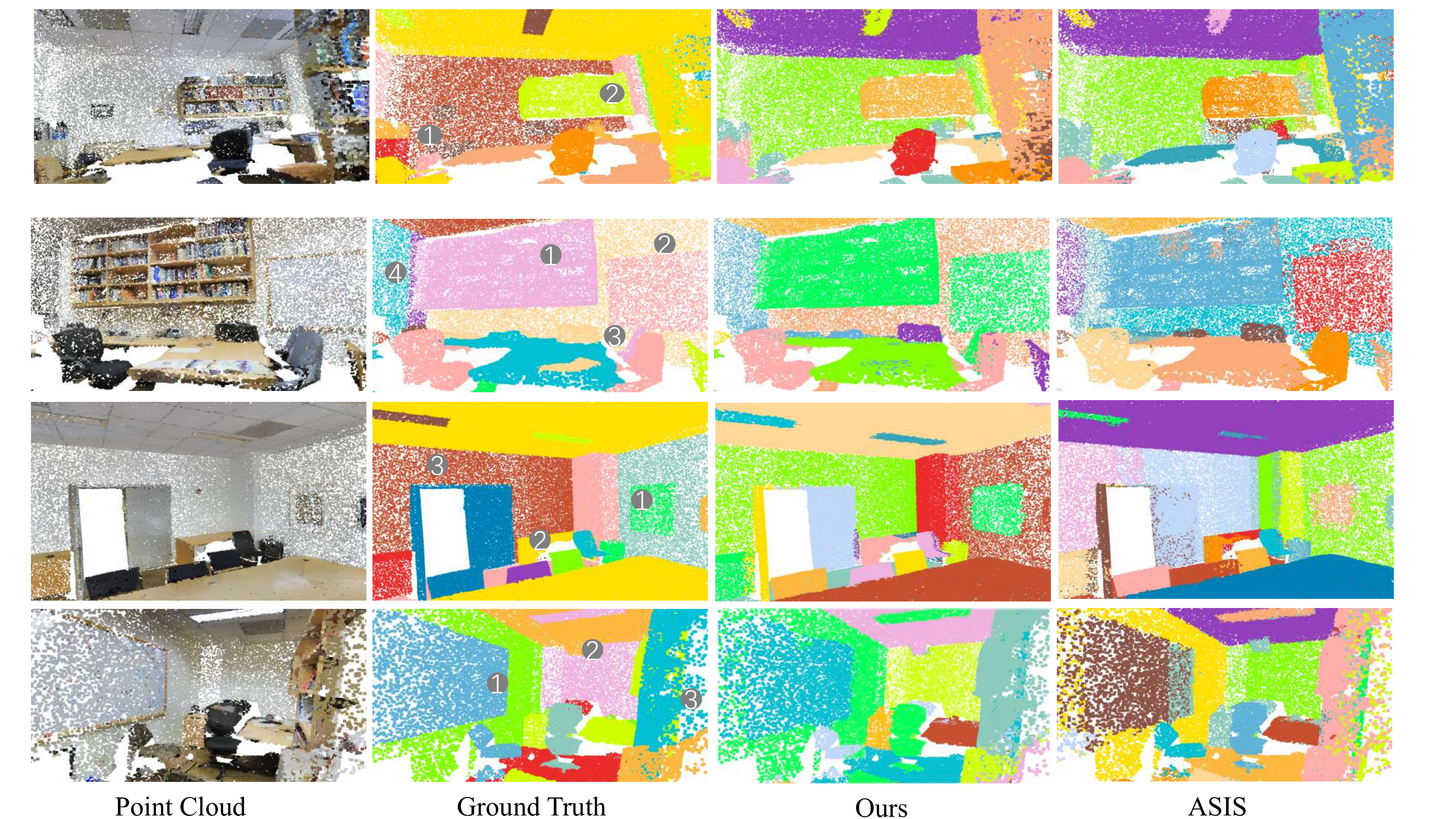}
		
	}
\end{center}
	\caption{\textbf{Qualitative results of our method on S3DIS dataset.} From left to right are: input point cloud, instance segmentation ground truth, the results of our method and the results of \cite{wang2019asis}. Note that different instance are shown with different colors, and the same instance are not necessarily have the same color in ground truth and prediction presentation.}
	\label{fig:results}
\end{figure*}

\subsubsection{Performance on S3DIS.}
We first compare the instance segmentation performance on both Area-5 and 6-fold. The results are presented in Table \ref{tab:s3dis_ins_results}. Our proposed MPNet achieve promising results and surpass the previous state-of-the-art approaches substantially by a large margin. The large improvement is mainly beneficial from the strong ability of the proposed prototypes memory. Qualitative results is show in Figure \ref{fig:results}.
In addition to instance segmentation, we also report the results of semantic segmentation and compare it with other methods. The performance is tested on all areas (6-fold), as shown in Table \ref{tab:seg_s3dis}. Although based on a simple PointNet++, we achieve even better quantitative results than other methods which are based on graph neural networks \cite{li2019deepgcns, wang2019dgcnn}.

\subsubsection{Performance on ScanNetV2.} 
In addition to S3DIS, we conduct experiments on ScanNetV2 \cite{dai2017scannet}. The instance segmentation results are reported in Table \ref{tab:scanet_val}, which is tested on the validation set. To make fair comparison, we select the methods that are based on PointNet or PointNet++. Our proposed MPNet outperforms previous methods over all overlap thresholds and dominant in many categories.

\begin{table*}[!htb]
		\caption{Instance segmentation results on ScannetV2 benchmark (validation set). Both results of mAP@0.25 and mAP@0.5 are reported. All methods except \cite{engelmann2018} are based on PointNet or PointNet++ (3D-BEVIS \cite{elich20193dbevis} is multi-view based method). }
	\begin{center}
	\setlength{\tabcolsep}{4.2pt} 
	\addtolength{\tabcolsep}{-1.5pt}
	\scriptsize
	\begin{tabular}{c|c|cc|cccccccccccccccccc}
		\toprule
		\multicolumn{1}{c|}{Method}  &\multicolumn{1}{c|}{Year} & \multicolumn{1}{c}{\begin{tabular}[c]{@{}c@{}}mAP\\ @0.25\end{tabular}} & \multicolumn{1}{c}{\begin{tabular}[c]{@{}c@{}}mAP\\ @0.5\end{tabular}} & \multicolumn{1}{|c}{bathtub} & \multicolumn{1}{c}{bed} & \multicolumn{1}{c}{shelf} & \multicolumn{1}{c}{cabinet} & \multicolumn{1}{c}{chair} & \multicolumn{1}{c}{counter} & \multicolumn{1}{c}{curtain} & \multicolumn{1}{c}{desk} & \multicolumn{1}{c}{door} & \multicolumn{1}{c}{other} & \multicolumn{1}{c}{picture} & refrig & shCur & sink & sofa & table & toilet & window \\
		\midrule
		MaskRCNN \cite{he2018maskrcnn} &2017 & 26.1 & 5.8 & 33.3 & 0.2 & 0.0  & 5.3 & 0.2  & 0.2 & 2.1 & 0.0& 4.5& 2.4& \textbf{23.8}& 6.5    & 0.0   & 1.4  & 10.7 & 2.0   & 11.0   & 0.6    \\
		SGPN \cite{wang2018sgpn} &2019 & 35.1& 14.3& 20.8& 39.0& \textbf{16.9}& 6.5& 27.5& 2.9& 6.9& 0.0& 8.7& 4.3& 1.4& 2.7& 0.0   & 11.2 & 35.1 & 16.8  & 43.8& 13.8   \\
		3D-BEVIS \cite{elich20193dbevis} &2019 & -& 24.8& 66.7& 56.6& 7.6& 3.5& 39.4& 2.7& 3.5& 9.8& 9.9& 3.0& 2.5& 9.8& 37.5  & 12.6 & \textbf{60.4} & 18.1  & 85.4   & 17.1   \\
		R-PointNet \cite{yi2018gspn} &2019& 40.0& 23.5& 51.3& 52.3& 12.5& 15.2& 61.8& 0.0& 1.5& 7.6& \textbf{29.0}& 11.7& 14.7& \textbf{25.0}   & 3.7   & \textbf{14.0} & 34.5 & 18.1  & 53.0   & 16.1   \\
		ASIS \cite{wang2019asis} &2019& 41.5& 24.0& 29.9& 50.5& 0.0& 16.7& 57.7& 0.0& \textbf{18.4}& 7.8& 14.8& 12.9& 1.8& 12.4   & 38.0  & 10.2 & 36.9 & 37.4  & 71.7   & 14.5   \\
		Ours& -& \textbf{49.3}& \textbf{31.0}& \textbf{69.4}& \textbf{59.8}& 2.7& \textbf{23.7}& \textbf{71.1}& \textbf{4.5}& 8.4& \textbf{18.3}& 11.6& \textbf{17.3}& 4.8& 21.8   & \textbf{57.0}  & 13.4 & 27.7 & \textbf{41.8}  & \textbf{87.3}   & \textbf{18.3}  \\
		\bottomrule
	\end{tabular}
\end{center}
\vspace{-1.5em}
	\label{tab:scanet_val}
\end{table*}

\begin{table}[htb]
	\caption{Inferencing time comparison on S3DIS Area-5 set. Forward time is network running time on GPU, whereas Postprocessing time is the BlockMerging algorithm introduced in \cite{wang2018sgpn}.
		ASIS is 45\% slower than our method in the forward process due to the usage of KNN, which is extremely time consuming. Reported time is running on a single 1080ti GPU with 4096 input points. }
	\small 
	\begin{center}
		\addtolength{\tabcolsep}{-3.5pt}
		\begin{tabular}{c|c|c|cc|c|c}
			\hline
			\hline
			\multirow{2}{*}{Method}  &\multirow{2}{*}{Backbone}    &   \multicolumn{3}{c|}{Inference Time (ms)} &  \multirow{2}{*}{mPre} &\multirow{2}{*}{mRec}\\
			\cline{3-5}
			& & Overall & Forward & {Post} & \\
			\hline
			SGPN\cite{wang2018sgpn} & PointNet & 730 & \textbf{22} &708 &36.0 &28.7 \\
			ASIS\cite{wang2019asis} &PointNet2 & 183 & 58 &\textbf{125} &55.3 & 42.4 \\
			Ours &PointNet2 &\textbf{165} &40 &\textbf{125} & \textbf{62.5} & \textbf{49.0} \\
			\hline
		\end{tabular}
	\end{center}

	\label{tab:s3dis_efficiency}
	\vspace{-2em}
\end{table}

\subsubsection{Speed Analysis.}
We compare the inference speed with other two methods: SGPN \cite{wang2018sgpn} and ASIS \cite{wang2019asis}. The whole evaluation process includes two parts: network forward  and instance grouping. The first part is to get per-point semantic labeling and instance embedding. The second part utilizes a grouping algorithm to find out instance groups. SGPN, which is based on PointNet, predicts a pair-wise affinity matrix to group points into instance clusters. Due to the large size of input point cloud, a huge memory is required. Meanwhile, as lots of heuristic parameters are introduced, the whole time for post-processing is much slower than our proposed method. Different from SGPN, ASIS utilize mean-shift for clustering embeddings to instance groups. Meanwhile, 
ASIS applies KNN for fusing semantic context from a fixed number of neighboring points, which is used on every input point. This operation is extremely time-consuming and fail to take fully advantage of computational resources. Compared with the above two approaches, our proposed MPNet is free from complex and time-consuming operations, showing the superiority in both effectiveness and efficiency.

\section{Conclusion}

In this paper, we propose a memory-augmented network to handle both category and pattern imbalance in point cloud instance segmentation. A memory module is introduced to alleviate the forgetting issue during the training process. The performance on the benchmarks shows the superiority of our method in both effectiveness and efficiency.

	\section{Appendix}
In this supplementary material, we provide more detailed experimental results, including: 
\begin{itemize}
	\item  Both qualitative and quantitative results on the ``Chair'' category in PartNet \cite{mo2019partnet};
\item  More visualization of our approach on S3DIS \cite{armeni2016s3dis} and ScanNetV2 \cite{dai2017scannet}.
\end{itemize}

\subsection{Experimental Results on PartNet dataset \cite{mo2019partnet}}
\begin{figure*}[htbp]
	\centering
	{
		\includegraphics[width=16cm, height=9cm]{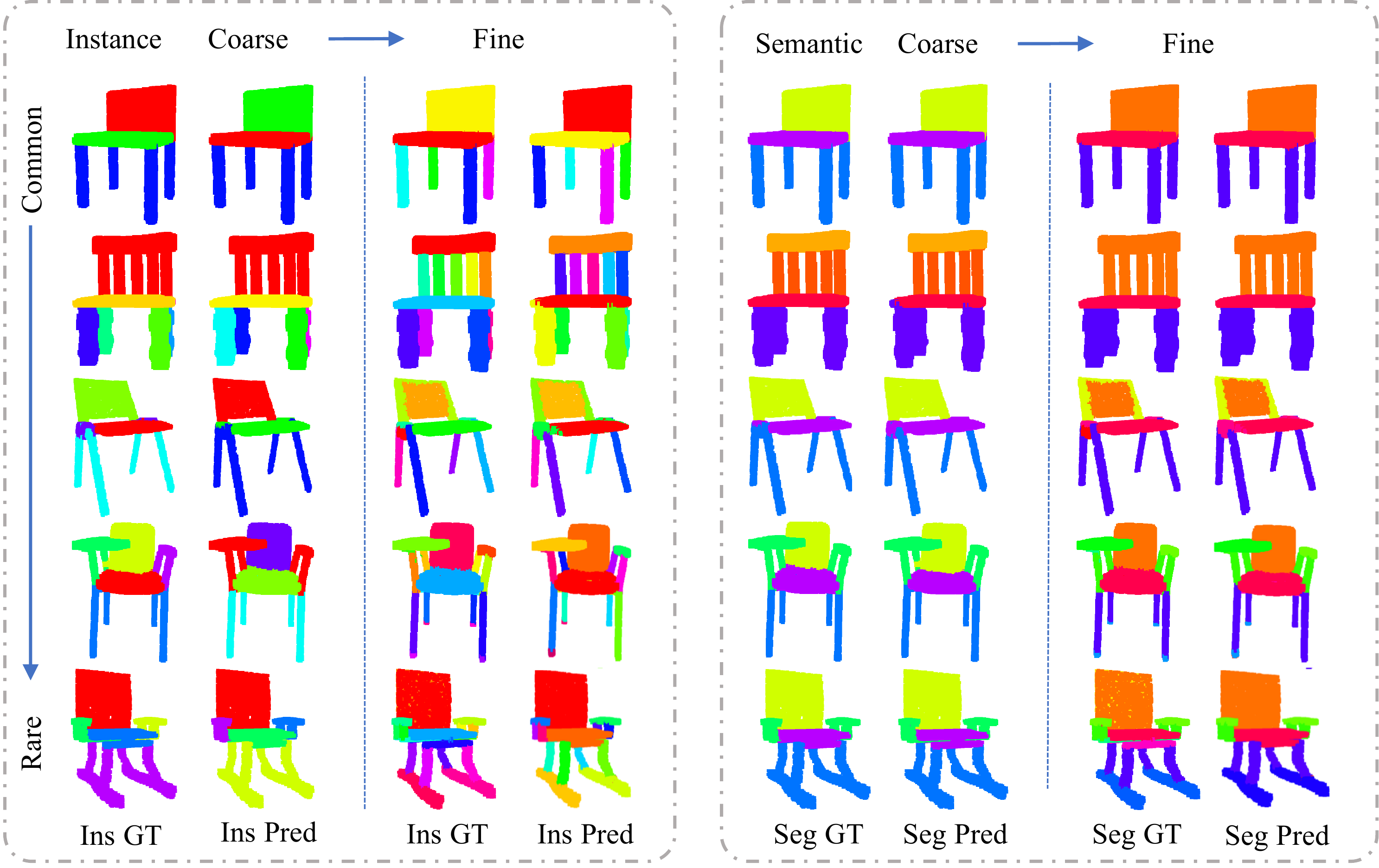}
	}
\vspace{-0.2cm}
	\caption{Visualization of the performance of on PartNet \cite{mo2019partnet}. Both coarse and fine-grained results are provided. Note that different instance are shown with different colors, and the same instance are not necessarily have the same
		color in ground truth and prediction presentation.}

	\label{fig:supply_partnet}
\end{figure*}

In Figure 3 in the main paper, to better understand the learned memory prototypes, we do visualization relying on the category of ``Chair'' in PartNet \cite{mo2019partnet}. 
PartNet \cite{mo2019partnet} is a consistent dataset of 3D objects with fine-grained and hierarchical 3D part annotations. 
In this section, we report the quantitative results in Table \ref{tab:parnet_chair}. \textbf{Level-1} refers to the coarsest annotation and \textbf{Level-3} refers to the most fine-grained annotation as defined in \cite{mo2019partnet}. 
For fair comparison, all results are evaluated with the same backbone PointNet++  \cite{qi2017pointnetplusplus}. Our method outperforms the previous methods by a large margin, showing the flexibility of our method to handle various types of input data. 
Moreover, visualization examples of the results are shown in Figure \ref{fig:supply_partnet}, indicating that our method can handle both rare and common cases well.

\begin{table}[htb]
	\caption{Comparison of the per-level performance of our method with the state-of-the-art methods on ``Chair'' category in PartNet \cite{mo2019partnet}. The performance is evaluated using part-category mAP, with IoU threshold of 0.5.
		All the results are achieved with the same backbone: PointNet++ \cite{qi2017pointnetplusplus}.  } 
	\begin{center}

		\begin{tabular}{c|c|c|c|c}
			\toprule[0.05cm]
			\textbf{Method} &Year &\textbf{Level-1} &\textbf{Level-2} &\textbf{Level-3} \\
			\toprule
			SGPN \cite{wang2018sgpn}&2019 &72.4 &25.4 &19.4 \\
			PartNet \cite{mo2019partnet} &2019 &74.4 &35.5 &29.0 \\
			GSPN \cite{yi2018gspn}&2019 &- &- & 26.8 \\
			Ours &- &\textbf{79.9} &\textbf{41.2} &\textbf{32.5} \\
			\bottomrule[0.05cm]

		\end{tabular}
	\end{center}
	
	\label{tab:parnet_chair}
\end{table}

\subsection{More Visualization Results}
\begin{figure*}[htbp]
	\centering
	{
		\includegraphics[width=15cm, height=8.5cm]{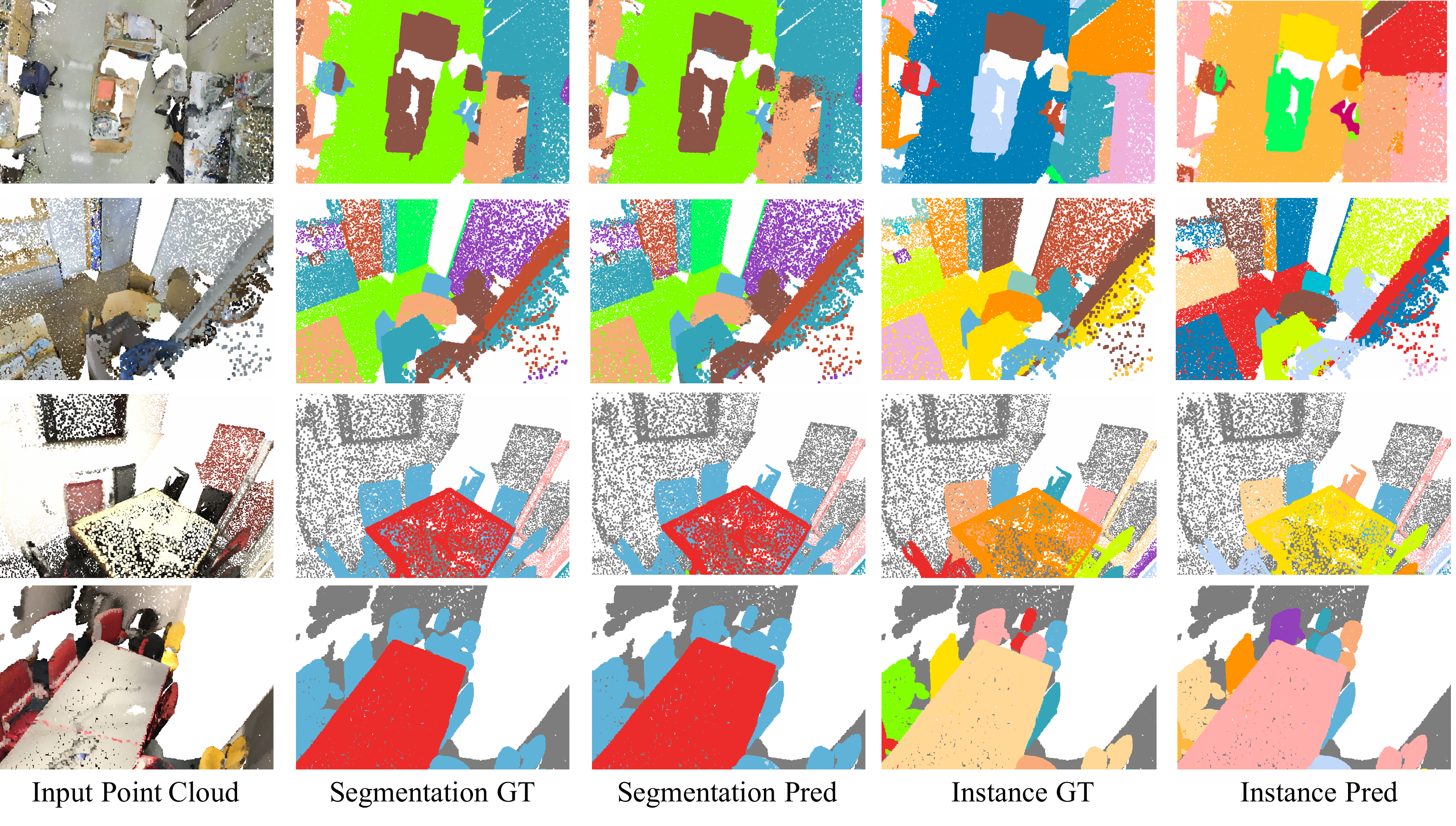}
	}
\vspace{-0.3cm}
	\caption{Visualization of the performance of on S3DIS \cite{armeni2016s3dis} and ScanNetV2 \cite{dai2017scannet}. Both instance and semantic segmentation results are provided. Note that different instance are shown with different colors, and the same instance are not necessarily have the same
color in ground truth and prediction presentation.}
	
	\label{fig:supply_result}
\end{figure*}

In the main paper, we illustrate the quantitative results on S3DIS \cite{armeni2016s3dis} and ScanNetV2 \cite{dai2017scannet} datasets in Table 2 and 4, respectively. 
Visualization examples of both semantic and instance segmentation results on S3DIS and ScanNetV2 datasets are shown in Figure \ref{fig:supply_result} in the following. 
	{\small
		\bibliographystyle{ieee_fullname}

	}
	
\end{document}